\documentclass[11pt,a4paper]{article}
\usepackage[hyperref]{emnlp2020}
\usepackage{times}
\usepackage{latexsym}

\usepackage{microtype}
\usepackage{graphicx}
\usepackage[ruled,vlined]{algorithm2e}
\aclfinalcopy 
\def\aclpaperid{3496} 

\title{Form2Seq : A Framework for Higher-Order Form Structure Extraction}

\author{Milan Aggarwal$^{1}$, Hiresh Gupta$^{2}$, Mausoom Sarkar$^{1}$, Balaji Krishnamurthy$^{1}$ \\ Media and Data Science Research Labs, Adobe$^{1}$ \\ Adobe Experience Cloud$^{2}$\\}

\begin{document}
\maketitle

\begin{abstract}
Document structure extraction has been a widely researched area for decades with recent works performing it as a semantic segmentation task over document images using fully-convolution networks. Such methods are limited by image resolution due to which they fail to disambiguate structures in dense regions which appear commonly in forms. To mitigate this, we propose Form2Seq, a novel sequence-to-sequence (Seq2Seq) inspired framework for structure extraction using text, with a specific focus on forms, which leverages relative spatial arrangement of structures. We discuss two tasks; 1) Classification of low-level constituent elements (TextBlock and empty fillable Widget) into ten types such as field captions, list items, and others; 2) Grouping lower-level elements into higher-order constructs, such as Text Fields, ChoiceFields and ChoiceGroups, used as information collection mechanism in forms. To achieve this, we arrange the constituent elements linearly in natural reading order, feed their spatial and textual representations to Seq2Seq framework, which sequentially outputs prediction of each element depending on the final task. We modify Seq2Seq for grouping task and discuss improvements obtained through cascaded end-to-end training of two tasks versus training in isolation. Experimental results show the effectiveness of our text-based approach achieving an accuracy of 90\% on classification task and an F1 of 75.82, 86.01, 61.63 on groups discussed above respectively, outperforming segmentation baselines. Further we show our framework achieves state of the results for table structure recognition on ICDAR 2013 dataset.  


\end{abstract}

\begin{figure*}[t]
\centering
\includegraphics[width=0.90\textwidth,height=2.55in]{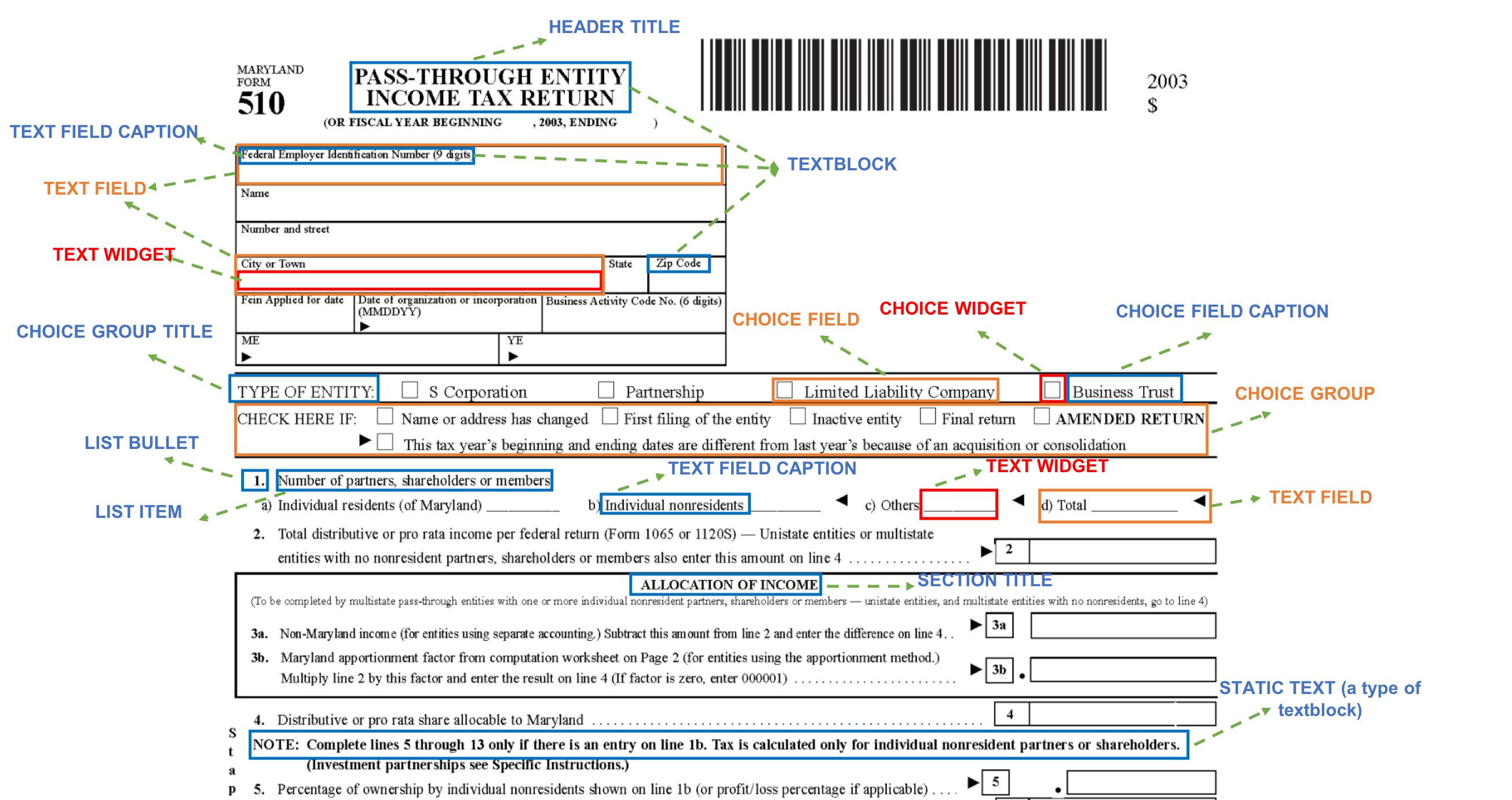} 
\caption{Different types of TextBlocks(Blue), Widgets(Red) \& higher order groups(orange) - ChoiceGroups, Choice Fields, Text Fields in a form. A text field comprises of 1) textblock(referred as text field caption) that describes what to fill \& 2) collection of widgets(text widgets). A choice group comprises of a title \& collection of choice fields. Textblocks \& widgets are classified into different types based on higher order group they are part of.}
\label{fig:describe}
\end{figure*}

\section{Introduction}

Various works \cite{tablerule, tableprice, historical, colorado} have studied semantic structure extraction for documents. Structure extraction is necessary for digitizing documents to make them re-flowable and index-able, which is useful in web-based services \cite{devices, mobile, dictionary,caseHTML}. In this work, we look at a complex class of documents i.e., Forms that are used to capture user data by organizations across various domains such as government services, finance, administration, and healthcare. Such industries that have been using paper or PDF forms would want to convert them into an appropriate digitized version \cite{caseHTML} (such as an HTML). Once these forms are made re-flowable, they can be made available across devices with different form factors\cite{devices,mobile}. This facilitates providing better form filling experiences and increases the ease of doing business since their users can interact with forms more conveniently and enables other capabilities like improved handling of filled data, applying validation checks on data filled in fields, consistent form design control\footnote{Please refer to supplementary for re-flow visualisation}.



To enable \textbf{dynamic rendering} of a form while \textbf{re-flowing} it, we need to extract its structure at multiple levels of hierarchy. We define TextBlock to be a logical block of self contained text. Widgets are spaces provided to fill information. Some low level elementary structures such as text and widgets can be extracted using auto-tagging capabilities of tools like Acrobat from the form PDF. However, such PDFs do not contain data about higher-order structures such as Text Fields, ChoiceGroups etc.

Document structure extraction has been studied extensively with recent works employing deep learning based fully convolution neural networks \cite{tableprice,historical,colorado} that perform semantic segmentation \cite{FCNSS,chen2014semantic,noh2015learning} on document image. Such methods perform well at extracting coarser structures but fail to extract closely spaced structures in form images (as discussed in the Experiments section). With increase in image resolution, number of activations(forward pass) and gradients(backward pass) increase at each network layer which requires more GPU memory during training. Since GPU memory is limited, they downscale the original image at the input layer which makes it difficult to disambiguate closely spaced structures, especially in dense regions(occurring commonly in forms) which leads to merging.

Figure \ref{fig:describe} shows different types of TextBlocks, Widgets and higher order groups. Given text blocks and widgets as input, our Form2Seq framework classifies them between different type categories. We hypothesize that type classification of lower level elements can provide useful cues for extracting higher order constructs which are comprised of such smaller elements. We establish our hypothesis for the task of extracting ChoiceGroups, Text Fields and Choice Fields. A Text Field is composed of textblock(textual caption) and associated widgets, as shown in figure \ref{fig:describe}. A choice group is a collection of boolean fields called choice fields with an optional title text (choice group title) that describes instructions regarding filling it. We study fillable constructs as they are intrinsic and unique to forms and contain diverse elementary structures.


The spatial arrangement of lower level elements with respect to other elements in a form are correlated according to the type of construct. For instance, a list item usually follows a bullet in the reading order; field widgets are located near the field caption. Similarly, elements that are part of same higher-order group tend to be arranged in a spatially co-located manner. To leverage this in our Form2Seq framework, we perform a bottom up approach where we first classify lower level elements into different types. We arrange these elements in natural reading order to obtain a linear sequence. This sequence is fed to Seq2Seq \cite{sutskever2014sequence} where each element's text and spatial representation is passed through a BiLSTM. The output of BiLSTM for each element is sequentially given as input to an LSTM \cite{hochreiter1997long} based decoder which is trained to predict the category type. For grouping task, we modify the framework to predict id of the group each lower level element is part of. Here the model is trained to predict same group id for elements that are part of same group. Our contributions can be listed as:

\begin{itemize}
    \item We propose Form2Seq framework for forms structure extraction, specifically for the tasks of element type classification and higher order group extraction.
    \item We show effectiveness of end-to-end training of both tasks through our proposed framework over performing group extraction alone. 
    \item We perform ablations to establish role of text in improving performance on both tasks. Our approach outperforms image segmentation baselines.
    \item Further, we perform table structure recognition by grouping table text into rows and columns achieving state of the art results on ICDAR 2013 dataset.
\end{itemize}


\section{Related Work}
Earlier works for document layout analysis have mostly been rule based relying on hand crafted features for extracting coarser structures such as graphics and text paragraphs \cite{rule1}. Approaches like connected components and others, were also used for extracting text areas\cite{rule3} and physical layouts\cite{rule4}. These approaches can be classified into top-down \cite{rule2} or bottom-up \cite{rule5}. The bottom-up methods focus on extracting text-lines and aggregating them into paragraphs. Top-down approaches detect layout by subdividing the page into blocks and columns. 


With the advancement in deep learning, recent approaches have mostly been fully convolution neural network (FCN) based that eliminate need of designing complex heuristics \cite{colorado,tableprice,historical}. FCNs were successfully trained for semantic segmentation \cite{FCNSS} which has now become a common technique for page segmentation. The high level feature representations make FCN effective for pixel-wise prediction. FCN has been used to locate/recognize handwritten annotations, particularly in historical documents \cite{handwriting}. \citeauthor{curtis} proposed a model that jointly learns handwritten text detection and recognition using a region proposal network that detects text start positions and a line follow module which incrementally predicts the text line that should be subsequently used for reading.

Several methods have addressed regions in documents other than text such as tables, figures etc. Initial deep learning work that achieved success in table detection relied on selecting table like regions on basis of loose rules which are subsequently filtered by a CNN \cite{tablerule}. \citeauthor{tableprice} proposed multi-scale, multi-task FCN comprising of two branches to detect contours in addition to page segmentation output that included tables. They additionally use CRF (Conditional Random Field) to make the segmented output smoother. However, segmentation based methods fail to disambiguate closely spaced structures in form images due to resolution limitations as discussed in experiments section. \citeauthor{gralinski2020kleister} introduced the new task of recognising only useful entities in long documents on two new datasets. FUNSD \cite{jaume2019funsd} is a small-scale dataset for form understanding comprising of 200 annotated forms. In comparison, our Forms Dataset is much larger having richer set of annotations. For task of figure extraction from scientific documents, \cite{figures} introduced a large scale dataset comprising of 5.5 million document labels. They find bounding boxes for figures in PDF by training Overfeat \cite{overfeat} on image embeddings generated using ResNet-101.



Few works have explored alternate input modalities such as text for other document related tasks. Extracting pre-defined and commonly occurring named entities from invoices like documents(using text and box coordinates) has been the main focus for some prior works \cite{katti2018chargrid,liu2019graph,denk2019bertgrid,majumder2020representation}. Text and document layouts have been used for learning BERT \cite{devlin-etal-2019-bert} like representations through pre-training and then combined with image features for information extraction from documents \cite{xu2020layoutlm,garncarek2020lambert}. However, our work focuses on extracting a much more generic, diverse, complex, dense, and hierarchical document structure from Forms. Document classification is a partly related problem that has been studied using CNN-only approaches for document verification \cite{sicre2017identity}. \citeauthor{HAN} have designed HAN which hierarchically builds sentence embeddings and then document representation using multi-level attention mechanism. Other works explored multi-modal approaches, using MobileNet \cite{mobilenets} and FastText \cite{fasttext} to extract visual and text features respectively, which are combined in different ways (such as concatenation) for document classification \cite{docclassmulti}. In contrast, we tackle a different task of form layout extraction which requires recognising different structures.



\citeauthor{colorado} also proposed a multimodal FCN (MFCN) to segment figures, tables, lists etc. in addition to paragraphs from documents. They concatenate a text embedding map to feature volume. We consider image based semantic segmentation approaches as baselines for the tasks proposed. We compare the performance of our approach with 1) their FCN based method and 2) DeepLabV3+ \cite{deeplabv3plus2018}, which is state of the art deep learning model for semantic segmentation.


\begin{figure*}[t]
\centering
\includegraphics[width=0.85\textwidth,height=2.53in]{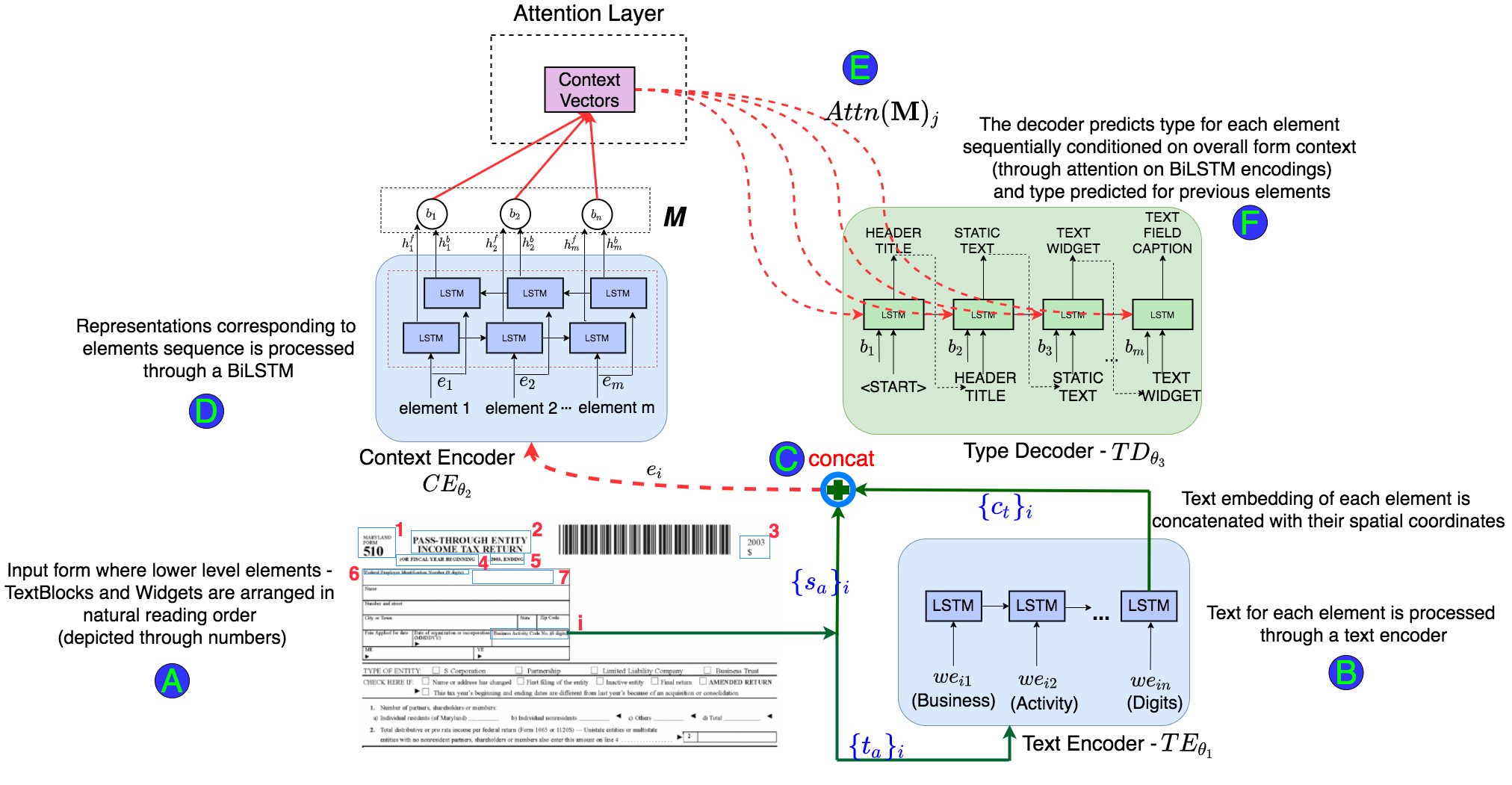} 
\caption{Model Architecture for element type classification. Different stages are annotated with letters.}
\label{fig:type}
\end{figure*}

\section{Methodology}


The spatial arrangement of a lower element among its neighbouring elements is dependent on the class of element. For instance, a list item usually follows a bullet in the reading order. Similarly, elements that are part of the same higher-order group tend to be arranged in a spatially co-located pattern. To leverage relative spatial arrangement of all elements in a form together, we arrange them according to a natural reading order (left to right and top to bottom arrangement), encode their context aware representations sequentially using text and spatial coordinates and use them for prediction. For each task, the decoder predicts the output for each element sequentially, conditioning it on the outputs of elements before it in the sequence in an auto-regressive manner (just like sentence generation in NLP). For group extraction task, our model assigns a group id to each element conditioning it on ids predicted for previous elements. This is essential to predict correct group id for current element (for instance, consider assigning same group id to elements that are part of same group).

Let a form be comprising of a list of TextBlocks ($f_{t}$) and list of widgets ($f_{w}$). We arrange $f_{e} = f_{t} \bigcup f_{w}$ according to natural reading order to obtain arranged sequence $a_e$ which is used as input for both the tasks (`A' in figure \ref{fig:type}).

\begin{figure*}[t]
\centering
\includegraphics[width=0.90\textwidth,height=2.53in]{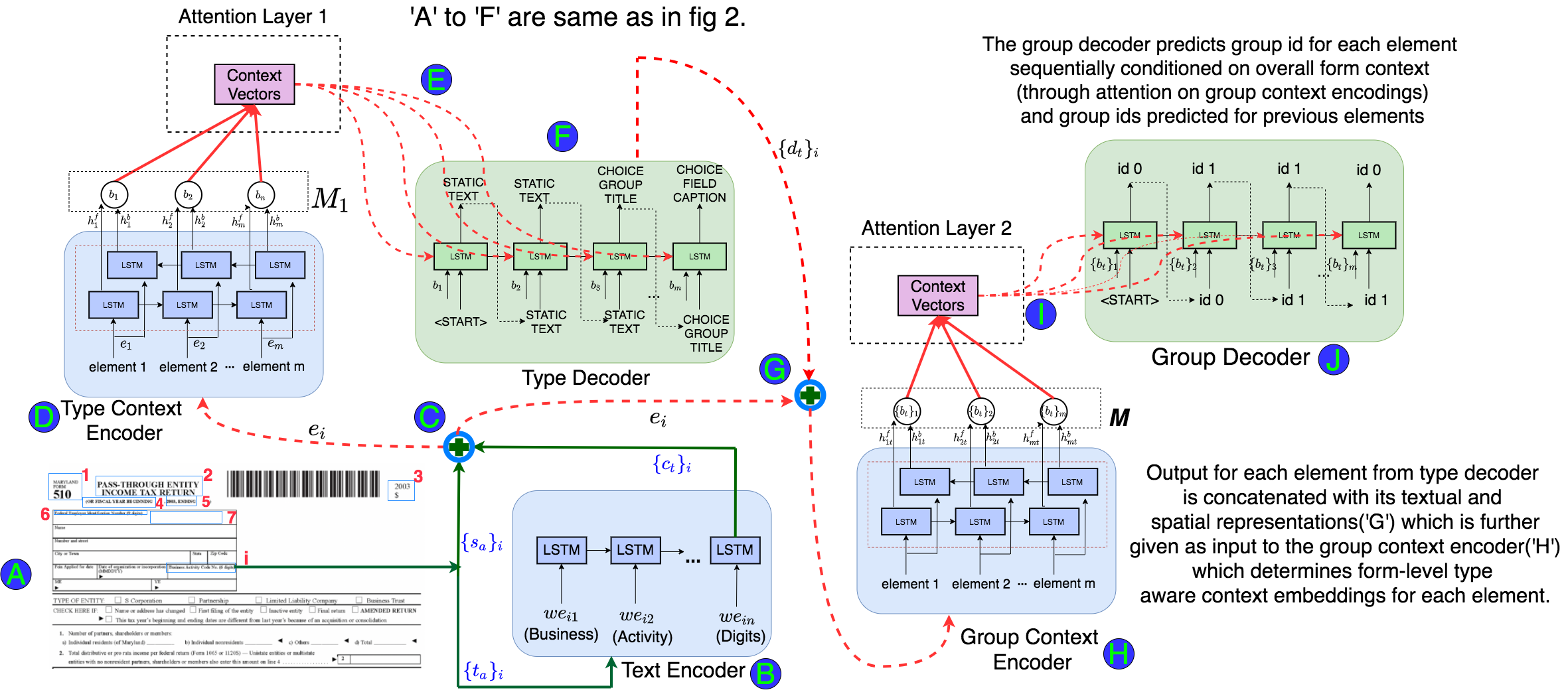} 
\caption{Architecture of our best performing model for group extraction leveraging type model shown in figure \ref{fig:type}.}
\label{fig:chgp}
\end{figure*}

\subsection{Element Type Classification}
Let $t_a$ and $s_a$ be the list of text content and spatial coordinates (x,y,w,h) corresponding to $a_e$, where x and y are pixel coordinates from top left corner in image and w \& h denote width and height of an element respectively. Our type classification model comprises of three sub-modules namely Text Encoder (TE) which encodes the text representation of each element, Context Encoder (CE) which produces context aware embedding for each element in the sequence, and Type Decoder (TD) which sequentially predicts type output. We discuss each of these modules in detail.

\noindent \textbf{Text Encoder :} Consider an element $\{a_e\}_i$ having text $\{t_a\}_i$ comprising of words $\{w_{i1},w_{i2}, ..., w_{in}\}$. Since the text information is obtained through PDF content, the words often contain noise, making use of standard word vectors difficult. To mitigate this, we obtain word embeddings using python library \textit{chars2vec}\footnote{https://github.com/IntuitionEngineeringTeam/chars2vec}. This gives a sequence of embeddings $\{we_{i1}, we_{i2}, ..., we_{in}\}$ which is given as input to an LSTM - $TE_{\theta_1}$, that processes the word embeddings such that the cell state $\{c_t\}_i$ after processing last word is used as text representation for $\{a_e\}_i$ (`B' in figure \ref{fig:type}). A widget's textual representation is taken as a vector of 0s.

\noindent \textbf{Context Encoder :} Consider a sequence element $\{a_e\}_i$ with corresponding textual representation $\{c_t\}_i$ and spatial coordinates $\{s_a\}_i$. These are concatenated (`C' in figure \ref{fig:type})) together to obtain $\{e\}_i$ representing the element. The sequence $e$ obtained is given as input to a BiLSTM - $CE_{\theta_2}$, which produces a context aware embedding $\{b\}_i$ for each element in the sequence (`D' in figure \ref{fig:type}). 

\noindent \textbf{Type Decoder :} The output from the previous stage is given as input to a final LSTM based decoder - $TD_{\theta_3}$, that sequentially outputs the category type for each element (`F' in figure \ref{fig:type}). Specifically, the decoder at time step i is given input $\{b\}_i$ to predict the type class of $i^{th}$ element. Additionally, we use Bahdnau attention mechanism \cite{bahdanau2014neural} to make $TD_{\theta_3}$ attend on context memory M (`E' in figure \ref{fig:type}) at each time step of decoding, where M is obtained by stacking $\{b_1;b_2; ..\}$ column-wise. This is to make it easier for decoder to focus on specific elements in sequence while predicting type for current element since elements sequence in a form tends to be very long. A linear layer with softmax activation is used over the decoder outputs for type classification.

We train all 3 modules - $TE_{\theta_1}$, $CE_{\theta_2}$ and $TD_{\theta_3}$ together using teacher forcing technique \cite{williams1989learning} and standard cross entropy loss.


\subsection{Higher Order Group Identification}
Our second task is to identify larger groups. Consider one such group - ChoiceGroup, comprising of a collection of TextBlocks and Widgets having different semantics(illustrated in figure \ref{fig:describe}). A ChoiceGroup contains 1) an optional choice group title which contains details and instructions regarding filling it; and 2) a collection of choice fields which are boolean fields such that each field comprises of a textual caption - choice field caption, and one or more choice field widgets. We formulate target label prediction for this task as that of predicting a cluster/group id for each element. Consider the element sequence $a_e$ such that elements $\{ \{a_e\}_{i1}, \{a_e\}_{i2}, ... \}$ are part of a group. We assign this group a unique number and train the model to predict same group number for each of these elements. Elements that are not part of any group are assigned a reserved group i.e. 0.

 We adopt a similar model as used for type classification except instead of type decoder, we have Group Decoder ($GD_{\theta_4}$) such that projection layer classifies each element into one of the groups. We hypothesize that category type of elements can be a useful clue for group decoder. To leverage the type information, we study a variant of our model - Cascaded Model, where we have a common text encoder but separate context encoders - $CE_{T}$ \& $CE_{G}$, and decoders - $TD$ \& $GD$, for the two tasks. Specifically, given a sequence of elements $a_e$ with combined textual and spatial representations $e$ (`C' in figure \ref{fig:chgp}), we first first feed them into type context encoder ($CE_{T}$, `D' in figure \ref{fig:chgp}) and type decoder ($TD$, `F' in figure \ref{fig:chgp}) as before to obtain decoder output sequence $d_t$ for each element. We modify the output types to categories which are relevant to the grouping task - ChoiceGroup Title, TextField Caption, ChoiceField Caption, ChoiceWidget, Text Widget, other TextBlocks. Since an element can be part of a field which is contained in choice group, we use two separate FC layers on decoder output to predict separate group ids for the element while determining choice groups and fields.

\begin{table*}[t]
 \centering
 \resizebox{\textwidth}{!}{%
 \begin{tabular}{cccccccccccc}

 \hline 
  Model & Choice & Text & Choice & Choice & TextField & Header & Section & Bullet & List & Static & Overall\\
   & Widget & Widget & GroupTitle & Caption & Caption & Title & Title & & Item & Text & \\
  \hline 

  $DLV3+$ & 68.24 & 96.66 & 57.90 & 76.28 & 86.10 & 83.55 & 55.43 & 48.89 & 75.94 & 69.37 & 84.18
  \\

  $MFCN$ & 0.0 & 81.25 & 0.0 & 0.0 & 46.87 & 69.42 & \textbf{71.47} & 90.03 & 54.26 & 11.29 & 48.59
  \\
  \hline
  $A_T$ (ours) & 67.77 & 85.92 & 56.18 & 66.81 & 80.72 & 82.38 & 57.20 & 82.70 & 81.84 & 70.24 & 76.92\\
   
  $B_T$ (ours) & 90.84 & \textbf{98.26} & 76.81 & 89.55 & \textbf{91.36} & 83.28 & 57.02 & 91.58 & \textbf{90.91 }& 82.18 & 88.87 \\
   
  $C_T$ (ours) & \textbf{91.83} & 96.89 & \textbf{78.93} & \textbf{90.53} & 91.27 & \textbf{85.88} & 67.48 & \textbf{93.55} & 90.78 & \textbf{85.31} & \textbf{90.06}
  \\
  
  \hline
\end{tabular}}

\caption{Element type classification accuracy of different ablation methods and baselines. Here $A_T$, $B_T$ and $C_T$ are different Form2Seq variants. $A_T$ gets only element's spatial coordinates as input, $B_T$ gets additional single bit depicting if an element is a TextBlock or a Widget in addition to their spatial coordinates, and $C_T$ gets both textual and spatial information as inputs but does not receive the additional bits provided to $B_T$. }
\label{typeResults}
\end{table*}

$TD$ outputs are concatenated with $e$ for each element (`G' in figure \ref{fig:chgp}) and given as input to group context encoder $CE_{G}$ to obtain contextual outputs sequence $b_t$ (`H' in figure \ref{fig:chgp}). The group decoder $GD$ (`J' in figure \ref{fig:chgp}) uses the sequence $b_t$ as input and attention memory (`I' in figure \ref{fig:chgp}) during decoding. For $d_t$, we purposely use outputs of type decoder LSTM and not final type projection layer outputs as determined empirically in experiments section. All five modules - $TE$, $CE_{T}$, $TD$, $CE_{G}$ and $GD$ are trained end-to-end for both tasks simultaneously.

\section{Experiments}

\subsection{Dataset}
\noindent \textit{\textbf{Forms Dataset:}} We have used our Forms Dataset comprising of 23K forms\footnote{Due to legal issues, we cannot release entire dataset. However, the part we plan to release will be large comprising of rich annotations and representative of our entire diverse set. It will be made available at: \url{https://github.com/Form2Seq-Data/Dataset}} across different domains - automobile, insurance, finance, medical, government (court, military, administration). We employed annotators to mark bounding box of higher order structures in form images as well as lower level constituent elements for each structure. There were multiple rounds of review where we suggested specific cases for each structure and patterns for correction to annotators. We discuss distribution of different structures across (train/test) splits for 10 element types : TextField Caption (129k/31.6k), TextField Widget (222k/533k), Choice Field Caption (35k/8.9k), ChoiceField Widget (39.2k/9.94k), ChoiceGroup Title (8.92k/2.28k), Header Title (10.2k/2.57k), Section Title (28.5k/7.25k), Bullet (56.4k/14.2k), List Item (58.9k/14.7k), Static Text (241.k/61.2k). For higher order structures, distribution of text fields and choice fields is same as that for their captions while for choice groups it is (15.5k/1.76k). Each form was tagged by an annotator(both lower and higher-level structures) and then reviewed by some other annotator. In $\sim$85\% forms, no corrections were made but some minor corrections were made in the rest 15\% cases after review phase. 
\\
\noindent \textit{\textbf{ICDAR 2013:}}
We also evaluate our approach on the table structure recognition task on ICDAR 2013 dataset. It comprises of 156 tables from two splits - US and EU set. We extract the images from the pdfs and train our model to extract the table structure by grouping  table text into rows and columns. We divide 156 tables into a set of 125 tables for training and 31 for testing following the strategy employed by \cite{row_column_table} and compare the performance of our approach with them.


\subsection{Implementation Details}
For text encoder $TE$, we fix size of text in a TextBlock to maximum 200 words. We use chars2vec model which outputs 100 dimensional embedding for each word and fix LSTM size to 100. For type classification, we use a hidden size of 500 for both forward and backward LSTMs in $CE_{T}$ and a hidden size of 1000 for decoder $TD$ with size of attention layer kept at 500. We tune all hyper-parameters manually based on validation set performance. Final type projection layer classifies each element into one of 10 categories. For grouping task, both isolated and cascaded model have exactly same configuration for $CE_{G}$ and $GD$ as for type modules. For cascaded model, type projection layer classifies each element into relevant type categories as discussed in Methodology section. We train all models using Adam Optimizer \cite{kingma2014adam} at a learning rate of $1 \times 10^{-3}$ on a single Nvidia 1080Ti GPU. We determined and used largest batch size(=8) that fits memory.


 \begin{figure*}[t]
\centering
\includegraphics[width=0.78\textwidth,height=1.2in]{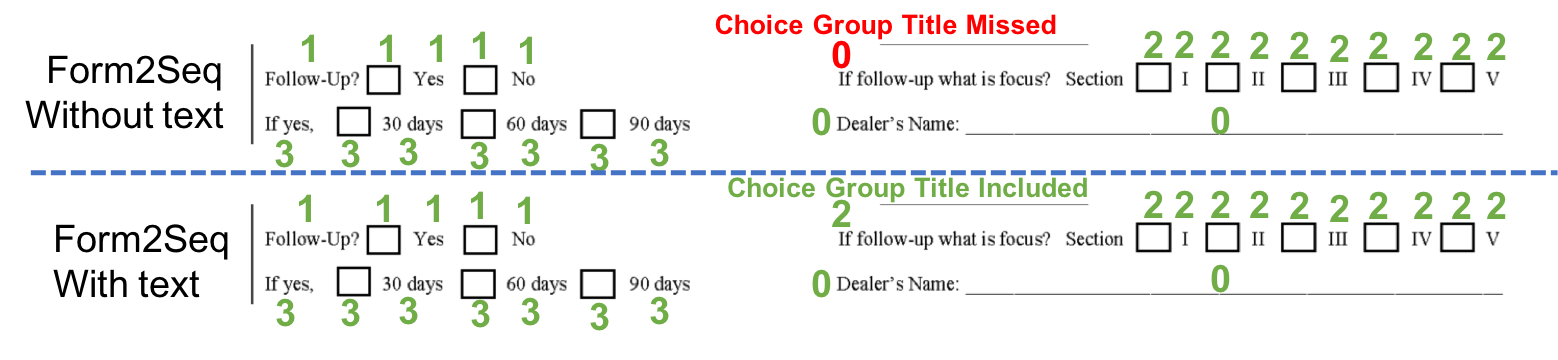} 
\caption{Predictions for a form snippet: Adding text input helps Form2Seq identify title which improves grouping.}
\label{fig:text_imp}
\end{figure*}

\subsection{Results and Discussion}
\noindent \textbf{Type Classification :}
  Results for type classification are summarized in Table \ref{typeResults}. We compare three models; $A_T$ - where we only give elements' spatial coordinates as input, $B_T$ -  where we additionally give  a single bit depicting if an element is a TexBlock or a Widget, and $C_T$ where both textual and spatial information is given as input. Using only coordinates yields inferior results since the model only has information regarding arrangement of elements. Adding textblock/widget flag significantly improves the overall accuracy by $\sim12\%$ ($A_T$ to $B_T$). Adding textual information (model $C_T$) improves the overall accuracy by $1.19\%$ to $90.06\%$. The accuracy for SectionTitle improves substantially from $57.02\%$ to $67.48\%$ and shows an improvement of $0.99\%$, $2.12\%$, $0.98\%$, $2.6\%$, $1.97\%$, $3.13\%$ for ChoiceWidget, ChoiceGroupTitle, ChoiceCaption, HeaderTitle, Bullet and StaticText respectively.

 \noindent \textbf{Group Identification :}
We report precision and recall numbers for the task of group extraction. Segmentation methods commonly use area overlap thresholds such as Intersection over Union(IoU) while matching expected and predicted structures(we evaluate baselines with an IoU threshold of 0.4). For our method, given a set of ground truth groups $\{g_1,g_2,g_3, ... ,g_m\}$ and a set of predicted groups $\{p_1,p_2,p_3, ... ,p_k\}$, we say a group $p_i$ matches $g_j$ iff the former contains exactly the same TextBlocks and Widgets as the latter. It takes into account all the lower elements which constitute the group (necessary to measure structure extraction performance). Thus, this metric is stricter than IoU based measures with any threshold since a group predicted by our method and evaluated to be correct implies that bounding box of prediction(obtained by taking the union of elements in it) will exactly overlap with expected group.


We first analyse the performance of our method on extracting choice groups. We consider different variants of our approach : 1) model $A_G$ - grouping in isolation; 2) model $B_G$ - both type and grouping task simultaneously with shared context encoder, type decoder attends on context encoder outputs while group decoder attends on context encoder outputs and type decoder outputs separately; 3) model $C_G$ - type identification trained separately, its classification outputs is given as input to group context encoder non-differentiably; 4) model $D_G$ - same as $B_G$ except separate context encoders for two tasks and softmax outputs concatenated with textual and spatial vectors as input to group context encoder; 5) model $E_G$ - same as $D_G$ except instead of softmax outputs, type decoder LSTM outputs are used; and 6) $F_G$(noText) - same as $E_G$ except spatial coordinates with isText signal used as input.

\begin{table}[h]
 \centering
 
 \begin{tabular}{cccc}

 \hline 
  Model & Recall & Precision & F-Score\\
  \hline 
  
  $DLV3+$ & 35.65 & 57.95 & 44.14 \\
  
  $MFCN$ & 16.97 & 11.86 & 13.96 \\
  \hline
  $A_G$(ours) & 51.18 & 55.48 & 53.24 \\
   
  $B_G$(ours) & 53.18 & 56.22 & 54.65 \\
   
  $C_G$(ours) & 55.9 & 57.15 & 56.51 \\
  
  $D_G$(ours) & 50.82 & 54.88 & 52.77 \\
  
  $E_G$(ours) & \textbf{58.67} & \textbf{60.81} & \textbf{59.72} \\
  
  $F_G$(ours) & 55.32 & 56 & 55.65 \\
  \hline
\end{tabular}
\caption{Comparison between F-scores of different models and baselines for ChoiceGroup Identification only. $A_G$ to $F_G$ are different variants of Form2Seq.}
\label{chgpResults}
\end{table}


\noindent Table \ref{chgpResults} shows joint training of both tasks improves F-score from $53.24$ to $54.65$ ($A_G$ to $B_G$) with improvement of $1.86$ if type information is incorporated non-differentiably($B_G$ to $C_G$). Our best performing model($E_G$) achieves an F-score of $59.72$. We observe that using type projection layer softmax outputs instead results in poor performance($E_G$ vs $D_G$). We observe that using \textbf{text} in Form2Seq($E_G$) performs $4.07$ points better in F-score vs. ablation $F_G$(w/o text). It can be seen in figure \ref{fig:text_imp} that $F_G$ misses choice group title(red), while Form2Seq with text($E_G$) extracts complete choice group\footnote{Please refer to supplementary for more visualisations}. \\

\begin{figure*}[t]
\centering
\includegraphics[width=0.92\textwidth,height=4.35in]{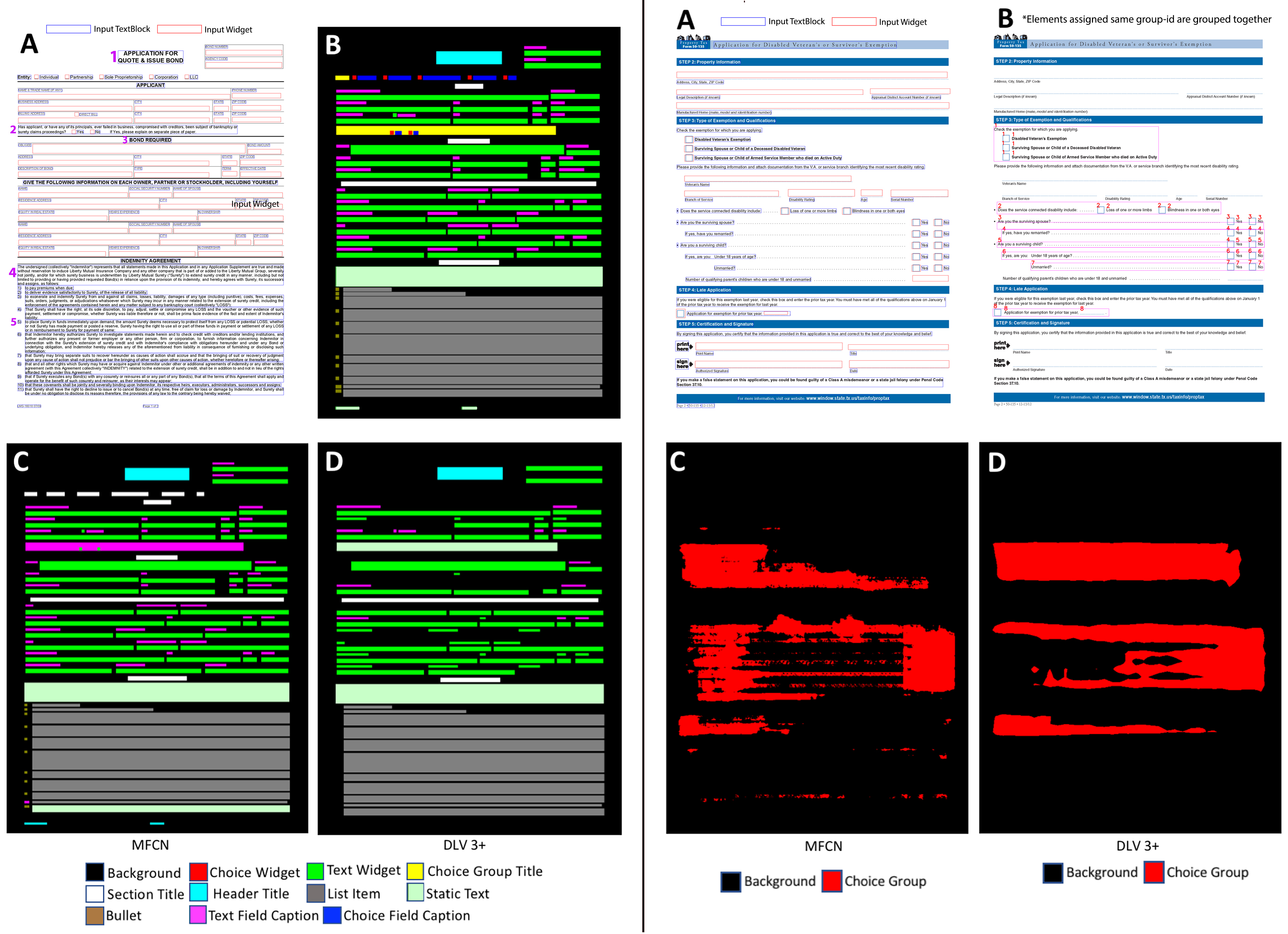} 
\caption{Examples of type classification (left) and choice group extraction (right). Top row shows form (A) and our outputs (B). For type predictions, we visualise our classification outputs as mask for understanding, and show post processed baseline outputs(through majority voting based on predicted masks). We can see that our Form2Seq framework makes better classifications for elements (2,3,5) marked in the top left image (1=Header Title, 2=Choice Group Title, 3=Section Title, 4=Static Text and 5=Bullet). For grouping task, elements highlighted with the same number by our model are predicted as part of same group(zoom in for viewing). Bottom row shows baseline segmentation outputs (C and D).}
\label{fig:visualization}
\end{figure*} 

\begin{table*}[t]
 \centering
 \begin{tabular}{c|ccc|ccc|cccc}
 \hline 
  \textbf{Construct} & \multicolumn{3}{c|}{\textbf{DLV3+}} & \multicolumn{3}{c|}{\textbf{MFCN}} & \multicolumn{3}{c}{\textbf{Ours}}   \\
  \hline
   & R & P & F & R & P & F & R & P & F  \\
  
  \hline
  Text Field & 43.64 & 34.63 & 38.62 & 37.12 & 38.94 & 38.0 & \textbf{71.59} & \textbf{80.6} & \textbf{75.82}  \\
  Choice Field & 61.93 & 44.42 & 51.73 & 31.45 & 14.24 & 19.6 & \textbf{83.48} & \textbf{88.71} & \textbf{86.01}  \\
  Choice Group & 43.25 & 53.5 & 47.83 & 30.99 & 26.85 & 28.77 & \textbf{59.27} & \textbf{64.2} & \textbf{61.63}  \\
  
  \hline
\end{tabular}
\caption{Recall(R), Precision(P) and F-score(F) of different methods on extracting different group structures together - text field, choice field and choice group simultaneously.}
 \label{chfieldResults}
\end{table*}

\begin{table*}[t]
 \centering
 \begin{tabular}{c|ccc|ccc|ccc}
Model & \multicolumn{3}{c|}{Table-Rows} & \multicolumn{3}{c|}{Table-Columns}    & \multicolumn{3}{c}{Average} \\
 \hline 
   & P & R & F1 & P & R &F1 & P & R & F1              \\ 
 \hline
  Baseline \cite{row_column_table} & \textbf{95.3} & 94.2 & 94.8 & 91.6 & 92.6 & 92.1 & 93.4 & 93.4 & 93.4 \\
  Ours & 94.2 & \textbf{96.1} & \textbf{95.1} & \textbf{95.7} & \textbf{92.9} & \textbf{94.3} & \textbf{95.0} & \textbf{94.5} & \textbf{94.7}
\end{tabular}
\caption{Comparison with baseline on Table Structure Recognition (identifying rows and columns) task on ICDAR-2013 dataset.}
\label{icdar-2013-comparison}
\end{table*}

\noindent \textbf{Comparison with baselines :}
 We consider two image semantic segmentation baselines - DeepLabV3+ (DLV3+) \cite{deeplabv3plus2018} and MFCN \cite{colorado}. For fair comparison, we implement two variants of each baseline - 1) only form image is given as input; 2) textblocks and widgets masks are given as prior inputs with image. We train both variants with an aspect ratio preserving resize of form image to 792x792. For MFCN, loss for different classes are scaled according to pixel area as described in their work. To classify type of an element, we post process prediction masks for baselines by performing a majority voting among pixels contained inside it for that particular element. For MFCN, without prior variant performed better, unlike DLV3+. We report metrics corresponding to better variant. As can be seen in table \ref{typeResults}, our best performing model ($C_T$) significantly outperforms both baselines in accuracy. Our model performs better for almost all category types. We observe that DLV3+ and MFCN are not able to perform well for all type classes simultaneously - DLV3+ performs sub-optimally for ChoiceWidget, Bullet and StaticText while MFCN performs poorly for ChoiceWidget, ChoiceGroup Title, Choice Field Caption even after loss scaling. We believe since forms are dense, such methods fail to distinguish different regions and capture complex concepts, for instance MFCN predicts `2'(shown in figure \ref{fig:visualization} (left)) as text field caption instead of choice group title due to widgets present around it.
 
For baselines, we match expected groups with segmented outputs through IoU overlap, keeping a threshold (0.40) for determining correct match. Since higher order groups span across different lower elements boundaries, it is not possible to leverage them to refine group masks predicted by baselines. Our proposed model (evaluated with stricter measure) outperforms DLV3+ (better baseline) by $15.58$ in F-Score(as can be seen in Table \ref{chgpResults}), even though it has lesser parameters(31.2 million) than DLV3+(59.4 million). Further, our main model ($E_G$) when evaluated through IoU overlap threshold of 0.40 achieves even higher recall, precision, F-Score of \textbf{74.3}, \textbf{78.6} and \textbf{76.3} respectively. Figure \ref{fig:visualization} shows outputs obtained using our approach and baseline methods. For grouping task (right), DLV3+ recognises couple of choice groups correctly but provides incomplete predictions in remaining regions, often merging them owing to its disability to disambiguate groups in dense areas. MFCN could not capture horizontal context between Choice group Title and Choice Fields and outputs broken predictions. In comparison, our model extracted 7 out of 8 choice groups correctly. \\


 \noindent \textbf{Extracting Higher Order Constructs Simultaneously:}
 We train our model to detect choice groups, text fields and choice fields together. To enable baseline methods to segment these hierarchical and overlapping structures simultaneously in separate masks, we use separate prediction heads on penultimate layer's output. Table \ref{chfieldResults} shows the results obtained. Our method works consistently well for all the structures outperforming the baselines. \\
 
\noindent \textbf{Table Structure Recognition : }
We further evaluate our proposed framework on a different task of grouping text in a table into rows and columns on publicly available ICDAR 2013 dataset. The input to our framework is the sequence of texts (arranged in natural reading order as usual) present in a table. We train our model to predict same group id for texts present in the same row and simultaneously detect columns in a similar manner using a separate prediction head. As a post processing step, we consider different sets of texts which are aligned vertically (sharing common horizontal span along the x-axis). We then consider the column group ids predicted by the model and assign majority column id (determined for a set using texts present in it) to all the texts in the set. The re-assigned ids are then used to determine different groups of texts to recognise columns. We perform similar processing while determining the final rows. \citeauthor{row_column_table} proposed to perform this task through constrained semantic segmentation achieving state-of-the-art results. Table \ref{icdar-2013-comparison} summarises the results obtained and compares our approach with \cite{row_column_table} showing our method obtains better F1 score for both rows, columns and average metrics (as used and reported in their paper).


\section{Conclusion}
We present an NLP based Form2Seq framework for form document structure extraction. Our proposed model uses only lower level elements - textblocks \& widgets without using visual modality. We discuss two tasks - element type classification and grouping into larger constructs. We establish improvement in performance through text info and joint training of two tasks. We show that our model performs better compared to current semantic segmentation approaches. Further we also perform table structure recognition (grouping texts present in a table into rows and columns) achieving state-of-the-art results. We are also releasing a part of our forms dataset to aid further research in this direction.




\bibliographystyle{acl_natbib}
\bibliography{emnlp2020}

\end{document}